\documentclass[10pt,twocolumn,letterpaper]{article}

\usepackage{cvpr}
\usepackage{times}
\usepackage{epsfig}
\usepackage{graphicx}
\usepackage{amsmath}
\usepackage{amssymb}
\usepackage{import}
\usepackage{subfigure}

\def \Rspace {\mathbb{R}}

\usepackage[pagebackref=true,breaklinks=true,letterpaper=true,colorlinks,bookmarks=false]{hyperref}

\cvprfinalcopy
\def\cvprPaperID{****} 

\ifcvprfinal\fi
\begin{document}

\title{Diagram Image Retrieval \\ using Sketch-Based Deep Learning and Transfer Learning}

\author{Manish Bhattarai$^{1,2,*}$, Diane Oyen$^{2}$, Juan Castorena$^{2}$, Liping Yang$^{1}$, and Brendt Wohlberg$^{2}$  \\
\\
        $^{1}$ University of New Mexico, Albuquerque, NM, USA  \\
          $^{2}$ Los Alamos National Laboratory, Los Alamos, NM, USA  \\
          $^{*}$ \textbf{Corresponding author}: Manish Bhattarai, ceodspspectrum@lanl.gov
}

\maketitle
\ifcvprfinal\thispagestyle{empty}\fi

\begin{abstract}
   Resolution of the complex problem of image retrieval for diagram images has yet to be reached. Deep learning methods continue to excel in the fields of object detection and image classification applied to natural imagery. However, the application of such methodologies applied to binary imagery remains limited due to lack of crucial features such as textures,color and intensity information. This paper presents a deep learning based method for image-based search for binary patent images by taking advantage of existing large natural image repositories for image search and sketch-based methods (Sketches are not identical to diagrams, but they do share some characteristics; for example, both imagery types are gray scale (binary), composed of contours, and are lacking in texture).
    We begin by using deep learning to generate sketches from natural images for image retrieval and  then train a second deep learning model on the sketches. We then use our small set of manually labeled patent diagram images via transfer learning to adapt the image search from sketches of natural images to diagrams. Our experiment results show the effectiveness of deep learning with transfer learning for detecting near-identical copies in patent images and querying similar images based on content.
\end{abstract}


\section{Introduction and motivation}

The patent industry involves the management and tracking of an enormous amount of data, much of which takes the form of scientific drawings, technical diagrams and hand sketched models. The comparison of figures across this dataset and subsequent retrieval based on similarity in real-time is extremely challenging \cite{mogharrebi2013retrieval}, \cite{vrochidis2012concept}, \cite{vrochidis2012patmedia}. 
We aim to track the spread of technical information by finding copies and modified copies of technical diagrams in patent databases and academic journals. Machine Learning (ML), and especially Deep Learning (DL) techniques offer the possibility of performing thousands of diagram/diagram comparisons across multiple databases in seconds. We present an ML approach that offers a high comparison accuracy with very little training data and, given a specific diagram and image of interest, under single shot and zero shot conditions can scan a database and retrieve all of the closest matches in that database for further review. 
Here, we present a deep learning approach that takes advantage of existing natural image repositories for image search and sketch-based methods applied to binary patent imagery.

The success of conventional CNN frameworks is widely acknowledged in image classification and cross-domain reconstruction applied to natural imagery when images have contextually rich information such as texture and pattern \cite{deng2009imagenet},\cite{zhou2003relevance}. These state-of-the-art frameworks fail when applied to diagrams, due to their contextually poor imagery. Standard One \cite{vinyals2016matching}, Zero-Shot(ZS) \cite{socher2013zero} and Few-Shot(FS) \cite{snell2017prototypical} techniques, originally developed for small datasets struggle to perform well on diagram-type imagery due to the domain variation and huge non-overlap in representation across these domains. 
Most technical diagrams, sketches and scientific drawings found in patents are binary images. They lack significant features such as texture, color and contrast. Also, there is structural variation because of rigid body transformations such as translation, rotation or perspective variations (i.e. viewpoint change). Classical image processing and computer vision tools such as key-point matching do not perform well given such transformations \cite{lecun2015deep}.  Typically, DL performance is robust against such rigid body transformations when trained with data augmentation techniques \cite{shorten2019survey}. However, due to the lack of sufficiently labeled patent image data available, DL models can easily over-fit when trained on the small datasets typical of patent-related imagery. 

Domain generalization\cite{muandet2013domain} and domain adaptation \cite{pan2010domain} techniques are gaining popularity as methods to address this data gap. Domain generalization provides a method to generalize the trained model over a broader dataset. Here, we apply the concept of domain generalization by pre-training an unsupervised DL model on a large set of sketches generated from natural images. We aim to  achieve a generalized representation of the latent space with the edge maps and then project the target patent dataset to this domain-invariant representation where differences between training domains are minimized by incorporating the proper loss functions. We explore this method in both few-shot and zero-shot conditions where the model is able to generalize the matches and make similarity predictions based on a small subset of the dataset for training. The model learns to recognize unseen matching pairs based on knowledge acquired from training of labeled similarity pairs.


While most image retrieval methods and algorithms are designed around natural imagery, sketch-based retrieval \cite{liu2017deep} provides promise as a means to further image retrieval related to patents. Our methods further extend their approach through the following steps:
\begin{enumerate}
    \item We use deep learning to generate sketches from natural images (using existing natural image repositories for image retrieval/image search/ image comparison).
    \item The large dataset of sketches created in (1) is used in the training for image retrieval (because if the original natural images match, we assume the corresponding sketches will match as well).
    \item The unsupervised deep learning model is trained on the sketches dataset for domain generalization.
    \item We use transfer learning (our small labeled subset of data for image query is used at this stage) to complete the image retrieval task based on the model trained in (3).
\end{enumerate}
We show that even under zero-shot and one-shot conditions, this framework surpasses classical retrieval frameworks for retrieval of similar binary images.

\section{Related work}

The requirements of a patent image retrieval system include full-image, sub-image, category-based image, rotation-, scale- and affine-invariant image searches, real-time performance, scalability, on-line learning, and semantic level interpretation. Although the combined set of requirements present significant challenges, we aim to address most of them in our approach. 
 
Content-based Image Retrieval (CBIR) makes use of low level visual features such as color, edges, texture, and shape to represent and retrieve images \cite{antani2002survey,smeulders2000content,zhou2003relevance}. \textbf{Relational skeletons} \cite{huet2001relational}, consider features such as relational angle and relational position between lines. The use of line segments in representing the image makes this approach sensitive to rigid body transformations such as rotations, translations and scaling. The \textbf{Edge Orientation AutoCorrelogram (EOAC)} approach used in the US patent retrieval system PATSEEK \cite{tiwari2004patseek} claims to be insensitive to translation and scaling; however the approach is computationally expensive and the complexity grows with the feature vector size. The use of user-defined thresholds makes the approach scale variant. The \textbf{Contour Description Matrix (CDM)} approach \cite{zhiyuan2007outward} uses canny edge detection for extracting contour information followed by converting each edge point to a polar coordinate system. While this approach is invariant to rigid body transformations, the size of the CDM is dependent on image resolution and the resulting processes are inefficient both computationally and memory-wise. The \textbf{Adaptive Hierarchical Density Histogram (AHDH)} method \cite{sidiropoulos2011content} along with the retrieval framework PATMEDIA \cite{vrochidis2012patmedia} exploits both local and global content. It uses both content-based (i.e image-based) as well as concept-based (text-based) retrieval and claims joint retrieval using both text and image give better retrieval performance. The algorithm calculates the adaptive hierarchical density histogram by computing the density of black pixels on a white plane after reducing noise and normalizing at the pre-processing stage. The ADHD process is made to retrieve the images belonging to the same category in the database and fails to retrieve similar images belonging to a different category. Besides, one needs to also manually set two different thresholds to make the system scale invariant which contradicts the idea of scale invariance. \textbf{Fisher vectors} based patent retrieval \cite{csurka2011xrce} uses Fisher vectors \cite{perronnin2007fisher} to represent patent images as low level features. For a pair of images, a dot product of fisher vectors is computed to measure the similarity between them. Similar to the ADHD approach, this approach does categorical based retrieval instead of similarity based retrieval.

\section{Datasets}
\begin{figure}[htb]
  \includegraphics[width=\linewidth]{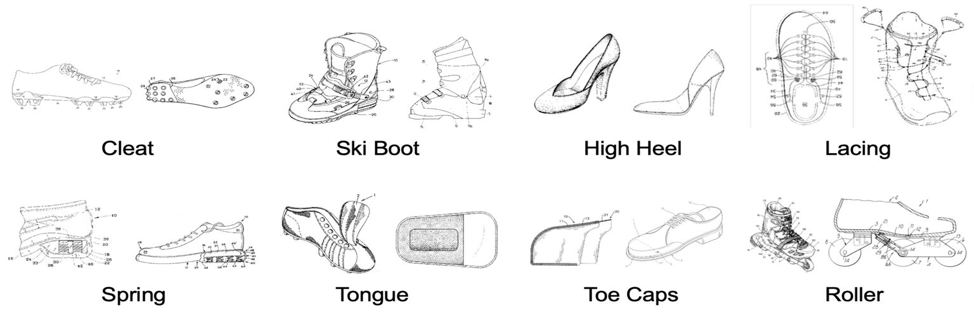}
  \caption{Two example images from each class of the Concept Dataset.}
  \label{fig:concept}
\end{figure}

The dataset used to train and test our model is taken from a patent image search benchmark \cite{vrochidis2012concept}. About 2000 sketch-type images are manually extracted from approximately 300 patents belonging to A43B and A63C IPC subclasses and contain types of foot-wear or portions thereof (henceforth termed "concepts"). The dataset consists of  8 concepts for this domain: cleat, ski boot, high heel, lacing closure, heel with spring, tongue, toe cap and roller blade. The details for the dataset can be found in \cite{vrochidis2012concept}.  The concepts dataset contains many dissimilarities within each class and is not suitable to train a classifier model to be used as a retrieval and matching framework. An example of the concepts dataset is shown in Figure~\ref{fig:concept}.
To ground-truth this concept dataset, we evaluated image similarity through manual pairwise comparisons made by three different non-experts and then determined a median out of all similarities. The pairwise similarity was quantified in the score range of 0-5 where, 5 - Same match, 4 - Slightly different, 3 - different perspective, 2 - sub-image, 1 - slightly different sub-image and 0 - dissimilar.

We used the UT Zappos50K shoe dataset \cite{yu2014fine} and the Generative Fashion dataset \cite{rostamzadeh2018fashion}  to generate the sketches for  domain generalization. The first dataset contains a total of 50K catalog images that were collected by Zappos.com while the second contains 293K high resolution fashion images. Using these two datasets, a retrieval performance was measured on the concepts dataset and fashion-MNIST \cite{xiao2017fashion} dataset respectively. The Fashion-MNIST dataset contains 70k images of gray scale fashion products in 10 categories.

\section{Methods}

Labeled benchmark datasets of natural images are easily accessible online, but labeled datasets of patent diagrams are more limited. 
To generate the sketches/edge-maps intended for usage as our custom shoe training dataset, we process the collection of natural images through the use of the Holistically Nested Convolutional Neural Nets(HCNN) \cite{xie2015holistically}. We train a Variational Auto-Encoder (VAE) \cite{kingma2014auto}, an unsupervised representation learning model, to approximate the distribution of the newly generated sketch dataset with a multi-dimensional Gaussian distribution with finite mean and variance. Once the model learns the representation of the data, we reuse this model via transfer learning on our small dataset for domain generalization \cite{oquab2014learning}. The idea of domain generalization is to learn from one or multiple training domains, to extract a domain-agnostic model which can be applied to an unseen domain.  We show  that, on passing the dataset through this learned model, it is able to achieve a minimal clustering of similar matches of the dataset. We augment this model with extra blocks of neural nets to construct a Siamese framework \cite{koch2015siamese} for fine tuning of the features on the latent space using triplet loss \cite{hoffer2015deep} that bring likely samples closer and push dissimilar samples farther away. During the training of the Siamese framework, the augmented block  is fine tuned with a small subset of the similarity matrix from the entire dataset. At the test phase, the samples that are used to query may/may not have been present during the training. If the similarity metric corresponding to the queried sample was used during training, then it is called Few-Shot/One-shot learning whereas if the no similarity metric corresponding to the queried sample was used during training, then it is called Zero-Shot learning. This applicability of one shot and zero shot retrieval with our framework relies on the knowledge gained during the domain generalization followed by intelligent fine tuning of the features. 
\begin{figure}[htb]
\centering
\includegraphics[width=\linewidth]{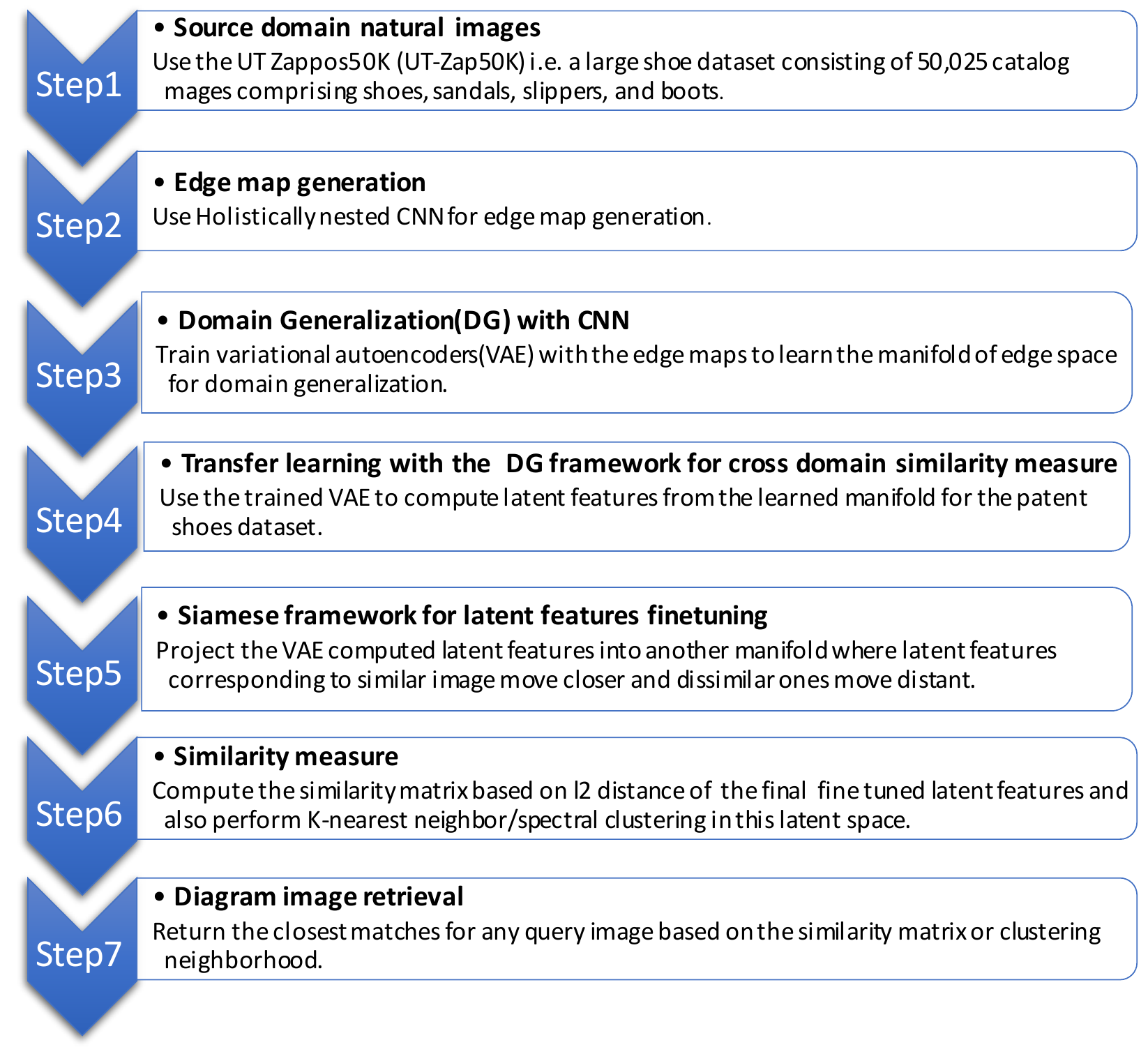}
 \caption{ Proposed model flow-chart }
\label{fig:flowchart}
\end{figure}

Once we have ideal clustering of the samples in the latent space via domain generalization and Siamese triplet loss based fine-tuning, we can use $k$-nearest neighbour($k$NN)  clustering  to return the more closely matched pairs. This could be incorporated into a retrieval tool to return the best set of matching images from the queried database.

To measure image similarity between two images $X,Y$ with corresponding pixels $\{x\}\in X$ and $\{y\}\in Y$ we use the mathematical expression:

\begin{align}
\begin{split}\label{eq:1}
    S(X,Y)={}& \sum_{x \in X} \sum_{y\in Y} K(x,y) 
\end{split}\\
\begin{split}\label{eq:2}
   ={}& \sum_{x \in X} \sum_{y\in Y} \phi(x)^T \phi(y) 
\end{split}\\
     ={}& \psi (X)^T \psi(Y) \label{eq:3}
\end{align}
where $K: \Rspace \rightarrow \Rspace$ is the operator denoting pixel similarity. Note that Eq. \ref{eq:1} is equivalent to the Kernel factorization of \cite{hofmann2008kernel} where image similarity is computed from features defined by the operator $\phi: \Rspace \rightarrow \Rspace$. Alternately, the similarity is a dot product between the transformed features as described by the function $\psi()$. These three expressions (1), (2) and (3) dictate the process of feature extraction, feature encoding and aggregation and database indexing respectively. 


\subsection{Holistically Nested Convolutional Neural Nets (HCN) for edge map generation}

\begin{figure*}
  \includegraphics[width=\textwidth]{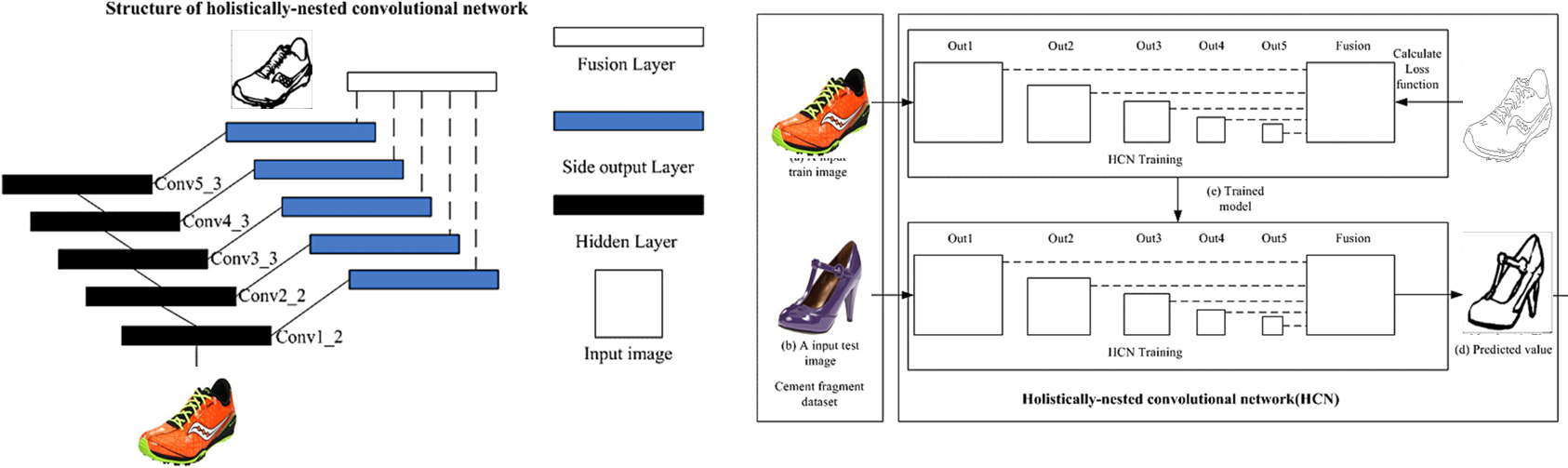}
  \caption{An overview of the Holistically Nested Edge Detection framework }
  \label{fig:HCNN}
\end{figure*}
 We exploit a state-of-the-art edge detection algorithm called Holistically Nested Convolutional Neural Nets (HCN) \cite{xie2015holistically} to generate the edge maps. This model utilizes an end-to-end  deep CNN framework for  image to image prediction where the input and output are natural image and edge map respectively. 
 
  The model comprises a modified VGG16 network \cite{krizhevsky2012imagenet} where the final pooling and fully connected layer is pruned. A deep supervision is established by connecting the side output layer to the last convolutional layer in each stage, Conv$1\_2$, Conv2$\_2$, Conv$3\_3$, Conv$4\_3$, and Conv$5\_3$, respectively. A convolutional layer with a kernel size of 1 is operated on the output of each of the previous layer outputs to compute side outputs  which are all then connected to a final fusion layer.   The framework is trained with image sketch pairs  and then tested with the shoes natural images. Figure \ref{fig:HCNN} demonstrates the HCNN framework for edge map generation. The overall loss function is given by 

\begin{equation}
    L(I,G,W,w)= L_{side} (I,G,W,w)+L_{fuse} (I,G,W,w)
\end{equation}
Where,\\
$L_{fuse}$  = fusion layer loss function, \\
$L_{side}$ = side output layer loss function, \\
$I$ =  raw input image, \\
$G$ =  ground truth binary segmentation map, \\
$W$ = collection of all other network layer parameters, \\
$w = {w(1),w(2),……..w(M)}$ : corresponding weights for each side output layer \\

\subsection{Domain generalization via Variational Auto Encoder(VAE)}
\begin{figure*}
    \includegraphics[width=\textwidth]{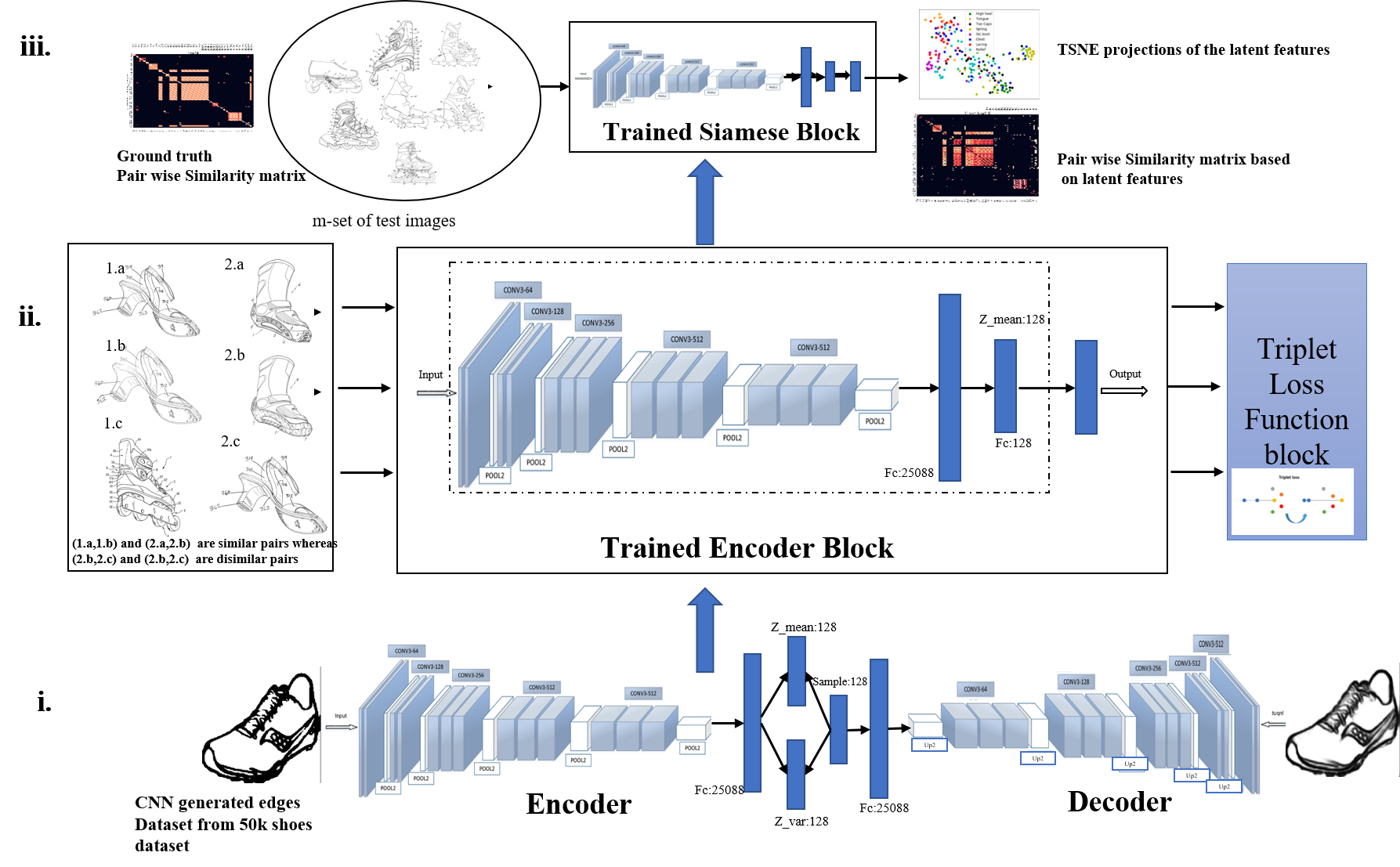}
    \caption{ An overview to the single-shot/zero-shot framework. \textbf{i.} Learning the latent representation of the 50K edge dataset with the Variational Auto-Encoder(VAE) for domain generalization.
\textbf{ii.}  Transfer learning of the trained Encoder block into the Siamese framework for training with concept dataset. 
\textbf{iii.}  t-distributed stochastic neighbor embedding(t-SNE) projection and similarity matrix computation based on latent features generated by the trained Siamese block.
}
    \label{fig:main_block}
\end{figure*}

\hspace*{-1cm}
\begin{figure*}
\centering
     \subfigure[]{\includegraphics[width=0.33\textwidth,page=1]{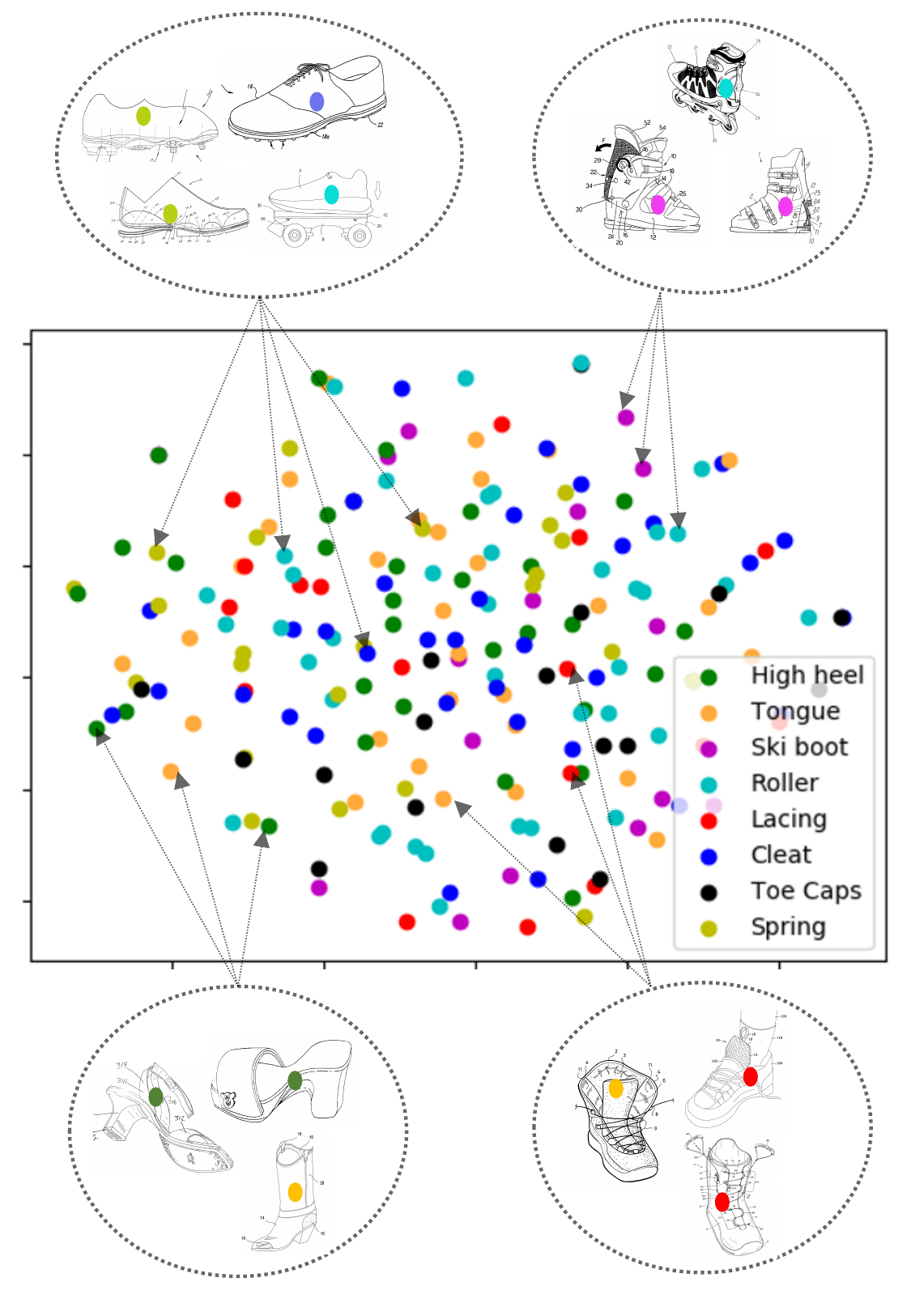}}
     \subfigure[]{\includegraphics[width=0.33\textwidth,page=2]{images/figure_TSNE.pdf}}
     \subfigure[]{\includegraphics[width=0.33\textwidth,page=3]{images/figure_TSNE.pdf}}
   
\caption{(a)TSNE Projection of image space \
(b)TSNE representation of VAE output feature space\ (c)TNSE representation of Siamese tuned output feature space}
\label{fig:TSNE}
\end{figure*}

We adopt the concept of domain generalization for representation learning. This refers to the learning representation of a domain dataset which makes it easier to extract significant information when building the matching framework. This kind of representation learning is usually done in unsupervised settings by leveraging the potential of the excess unlabeled dataset. Domain generalization tries to approximate the latent space/manifold over which the data of interest can be projected for categorization,clustering, or matching. 

We aim to combine a self supervised data representation achieved via a pre-training process and fine tune the model through transfer learning. We use a Variational Auto-Encoder (VAE) which is implemented on an explicit reconstruction loop that focuses on achieving per-pixel reconstruction. This VAE is trained for the purpose of unsupervised data representation and uses the encoder framework in a Siamese framework to achieve a benchmark performance of image matching/retrieval via transfer learning.

The VAE builds generative models of complex distributions of the sketched shoes dataset. It uses a CNN based function approximator to approximate an otherwise intractable function.  The encoder encodes the input data into the mean and variance statistics of the latent space and then samples data points from the Gaussian distribution computed from the statistics. The decoder tries to reconstruct the input data based on the sampled points. The framework trains in an end-to-end fashion where the objective for the encoder is to generate the statistical encoding in such a way that the difference between the input image and the reconstructed image are minimized. 

The VAE is incorporated to generate an observation x from some hidden variable z such that $p(z|x)$(intractable distribution) is approximated by another distribution $q(z|x)$ via approximate inference. With the objective to minimize the KL divergence between these two distributions $q(z|x)$ and $p(z|x)$ and also minimize the reconstruction error, we can write the overall loss function as  \\

\begin{equation}
    \theta(x) = KL(q_{\phi}(z|x)||p(z|x))+L(p_{\theta},q_{\theta}),
\end{equation} 
where
\begin{equation}
L(p_{\theta},q_{\theta})= E_{q_{\phi}(z|x)}[log p_{\theta}(x,z) - log q_{\theta}(z|x)]
\end{equation}
The first term represents the reconstruction error  reconstruction likelihood and the second term ensures that our learned distribution q is similar to the true prior distribution p.

\subsection{Single shot/zero shot training for image retrieval using a Siamese framework}

Once we have obtained the optimal latent representation of the latent space with the VAE, the encoder framework can be used to extract the optimal latent representation for our small dataset. To achieve a better cluster between similar image pairs, we implement triplet loss for a Siamese network. To compute triplet loss, we consider an anchor  i.e. a reference from which the distance will be calculated to a positive sample (e.g. sample with  a large paired similarity score) and negative sample (e.g. sample with 0 paired similarity score). If we consider the anchor, positive and negative sample images as ${x_i^a,x_i^p,x_i^n}$ and corresponding embedding vectors as ${f_i^a,f_i^p,f_i^n}$, then the triplet loss $l_{triplet}$ is given as \\

\begin{equation}
l_{triplet} = 1/N \sum_{i=0}^N max(0,||f_i^a-f_i^p||_2^2 - ||f_i^a-f_i^n||_2^2 + \alpha )
\end{equation}
Where $N$ is the batch size and $\alpha $ is a constant factor. 

Figure \ref{fig:main_block} gives a broader overview of the implemented methodology. In figure \ref{fig:TSNE}, we can see TSNE embedding corresponding to latent features at different stages of the framework.

\section{Experiments and results}

We used the Keras  framework with a Tensorflow backend to train and fine tune the proposed model. All experiments were conducted on the Darwin cluster at the Los Alamos National Labs. This cluster is equipped with Intel(R) Xeon(R) Gold 6138 CPUs and 8 GeForce RTX 2080Ti GPUs. The dataset for  training was constructed from the similarity matrix as triplets. The split was done 60\%(training+validation) and 40\%(test). A relatively larger test set was chosen to measure the performance of the model under One-Shot and Zero-Shot conditions. 

The implementation of the HCNN framework was borrowed from \footnote{\href{https://github.com/moabitcoin/holy-edge}{https://github.com/moabitcoin/holy-edge}}.  Considering the pertained model, the model used to generate the pseudo-sketches from the natural image datasets as discussed in the Dataset section. 

We constructed the deep VAE from the standard VGG16 architecture by trimming  the fully connected layer of VGG16 and augmenting the left model with a single layer VAE model to form an encoder and then combining with an equivalent decoder as shown in bottom of fig \ref{fig:main_block}. We validated the VAE model with the following hyper-parameters: a number of  convolutional layers and encoding dimensions for mean and variance. The  configuration including a 5-layered Convolutional block and 128 encoding dimensions achieved the best reconstruction accuracy on the pseudo sketch datasets. We used the batch size of 64, Adam optimizer and  learning rate of 0.001 for training the VAE. 

Next, the trained encoder model from the previous step was augmented with a fully connected block(FC) to construct a Siamese network whose triplet loss is computed from the set of three input images. We experimented with additional loss functions including the Contrastive loss \cite{hadsell2006dimensionality}, Constellation loss\cite{medela2019constellation}  and n-pair loss\cite{sohn2016improved}. Triplet loss provided the best similarity performance. We also investigated the performance of the Siamese framework by i) training the FC while freezing the encoder block and ii) training both FC and encoder in end-to-end fashion. We observed that the second approach achieves superior performance. This is due to a larger allowance for parameter learning for tuning the feature space.  
From figure \ref{fig:TSNE}, considering different groupings of similar items, the distance between sample points decrease in the feature space from a to c as shown by the TSNE\cite{maaten2008visualizing} projection. The data points are randomly distributed in the original pixel space and the pre-trained VAE achieved an improved level of closeness between similar samples without any knowledge of the patent dataset. Furthermore, with fine tuning utilizing the Siamese framework, the grouping of the samples based on their similarity is achieved.

\begin{figure}[htb]
  \includegraphics[width=\linewidth]{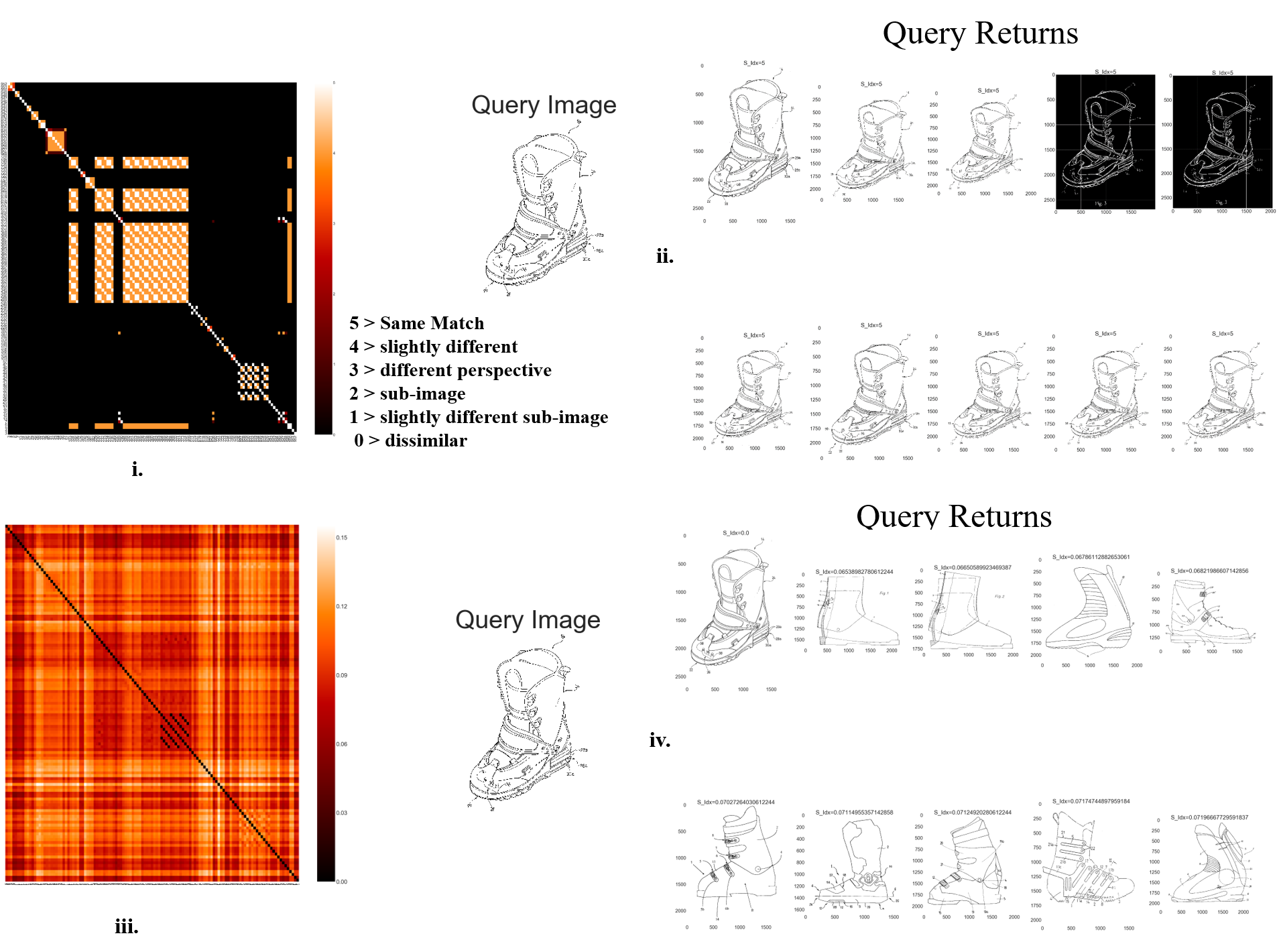}
  \caption{i) The pairwise ground truth similarity matrix  corresponding to the class ski boot. ii) Illustration of the database returns for the given query image. iii) Pairwise Similarity Matrix based on the Structural Similarity Index Measure (SSIM) between the images iv) Query returns based on the similarity matrix iii.}
  \label{fig:ssim_rtrv}
\end{figure}

\begin{figure}[htb]
  \centering
  \subfigure[Ground truth Similarity matrix]{\includegraphics[width=0.45\linewidth,height=4.2cm]{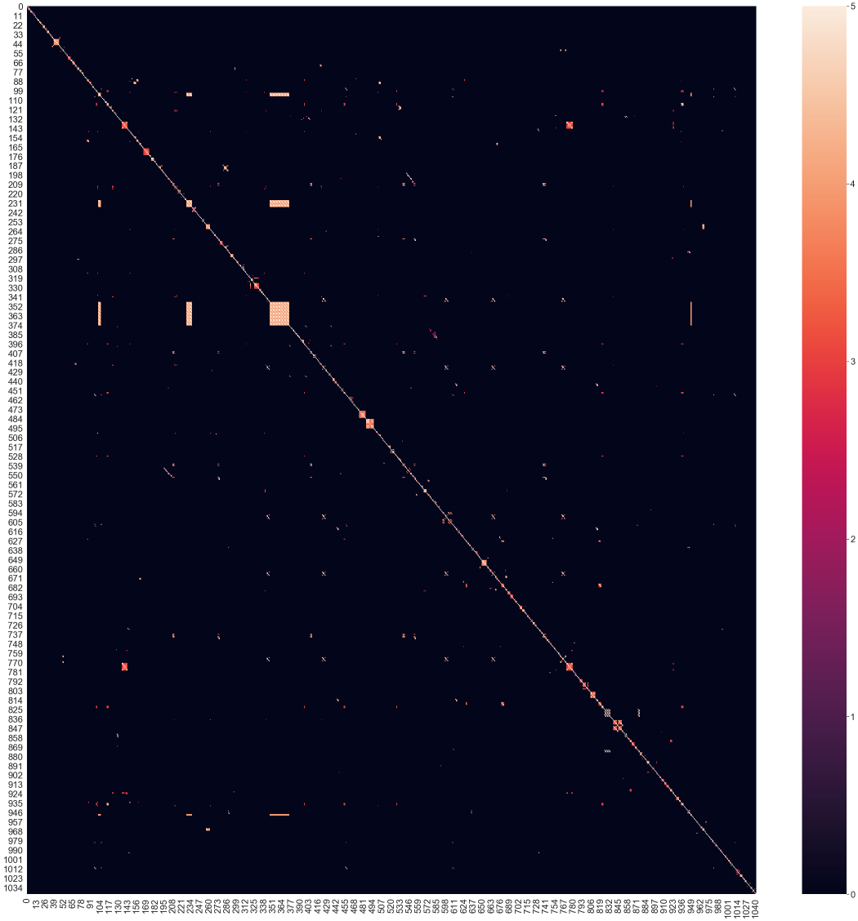}}
  \quad
  \subfigure[Predicted Similarity matrix(one-shot and zero-shot)]{\includegraphics[width=0.45\linewidth,height=4.2cm]{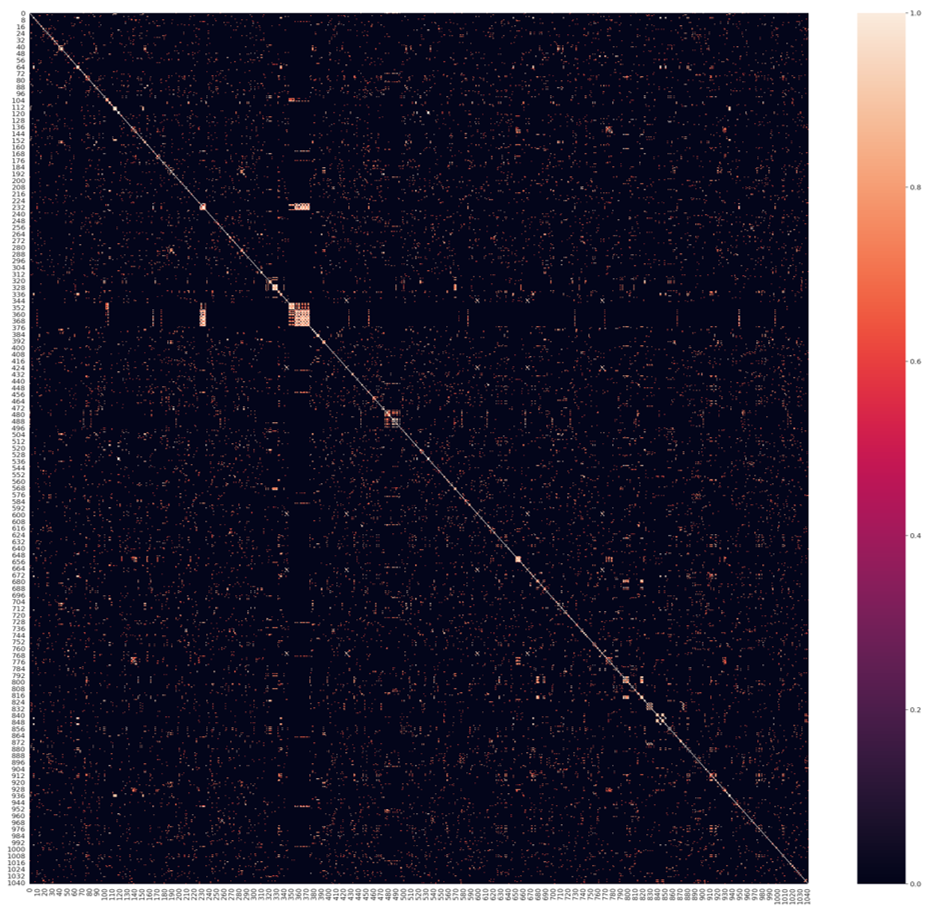}}
    
  \caption{Similarity matrix predictions. In ii) the similarity matrix is  predicted by the Siamese framework in  Few-Shot and Zero-Shot conditions. 
}
  \label{fig:Sim_matrix}
\end{figure}

\begin{figure}[htb]
  \centering
  \subfigure[Ground truth Similarity matrix based retrieval.]{\includegraphics[width=0.45\linewidth,height=0.36\linewidth]{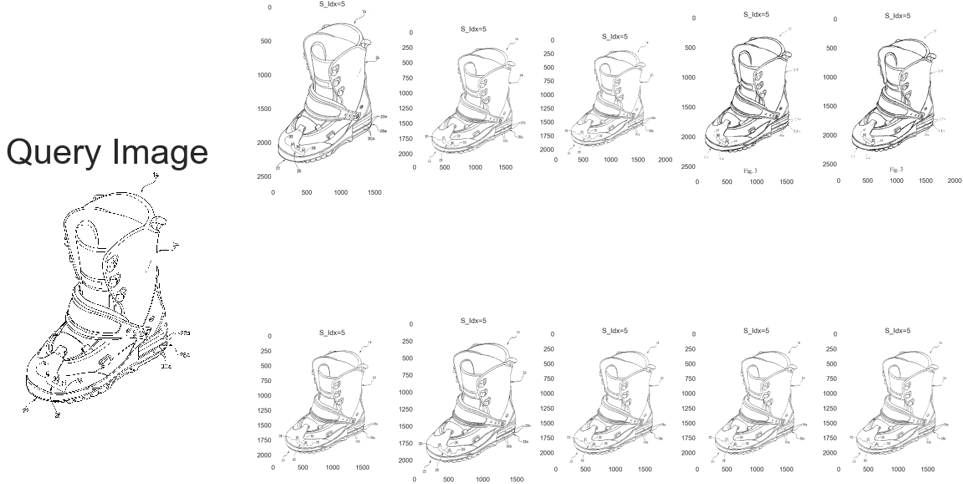}}
\quad
  \subfigure[Predicted Similarity matrix based retrieval(one shot).]{\includegraphics[width=0.45\linewidth,height=0.36\linewidth]{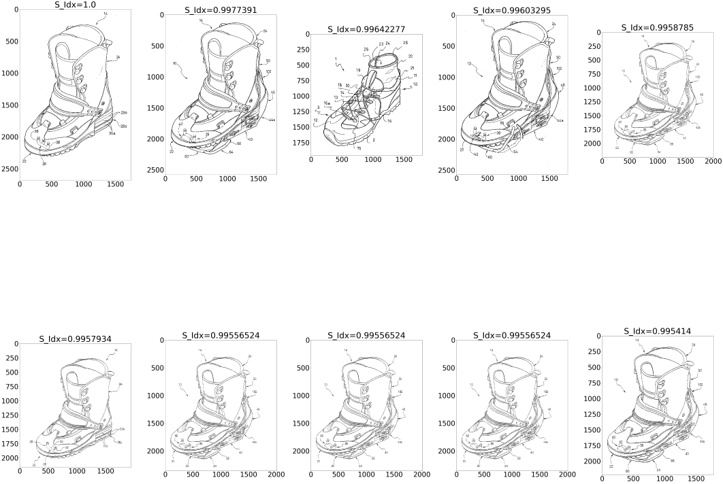}}
  
\subfigure[Ground truth Similarity matrix based retrieval.]{\includegraphics[width=0.45\linewidth,height=0.33\linewidth]{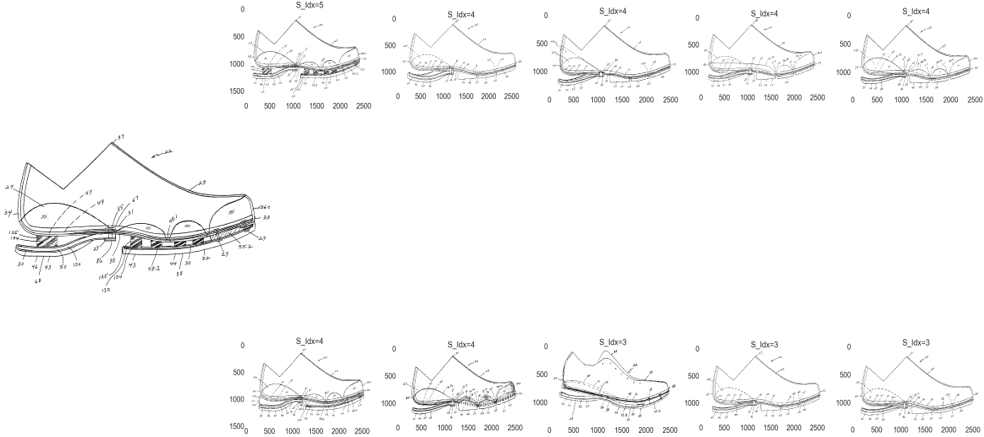}}
  \quad
 \subfigure[Predicted Similarity matrix based retrieval 
(zero shot).]{\includegraphics[width=0.45\linewidth,height=0.36\linewidth]{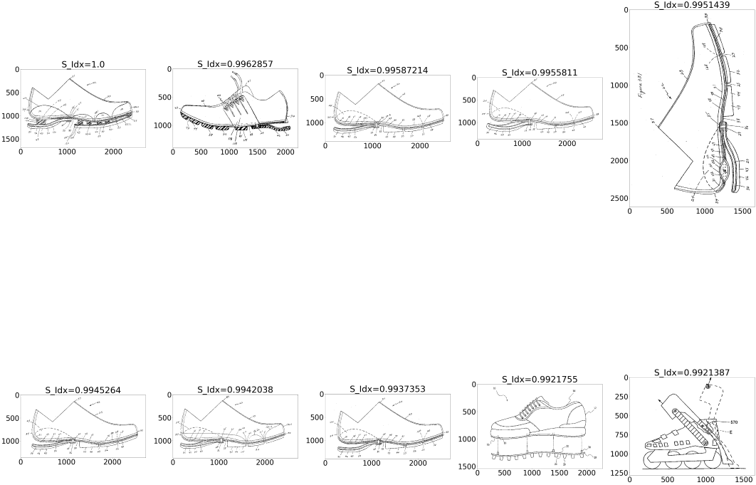}}
\caption{Predicted Similarity matrix based retrieval }
  \label{fig:retr_results}
\end{figure}

Before any retrieval task, we construct a full similarity matrix that encodes all computed pairwise similarities between the elements of the dataset. Pairwise similarities can be quantified as the cosine similarity or as the Euclidean distance between the features collected at the output of the Siamese layer followed by normalization of values to fall on a scale of 0-5.  For the patent shoes concept dataset, the similarity matrix is of the size $1042 x 1042$. To query any image from the row of the matrix, we process the corresponding columns and sort them based on the score (i.e highest to lowest score) and return the first $k$ elements of the sorted array. Figure \ref{fig:ssim_rtrv} is an example comparing ground truth and Structural Similarity Index Measure(SSIM) \cite{hore2010image} based similarity matrices for the retrieval of a queried image. 
However, if one needs to perform retrieval for additional images outside of the database, then the results of the query will be based on a newly computed similarity matrix that includes the newly added images. Figure \ref{fig:Sim_matrix} shows the ground truth similarity matrix and predicted similarity matrix based on both One-Shot and Zero-Shot conditions. 
Figure \ref{fig:retr_results} demonstrates the retrieval results under One-Shot and Zero-Shot conditions. 
To generalize our retrieval framework to a dataset with no available baseline similarities, we train using a binary similarity matrix with a score of 0 for intra-class and a score of 1 for inter-class images. This framework is likewise trained in zero-shot and one-shot conditions.  
The output features of the Siamese network in the test dataset were clustered with $k$-NN to obtain the nearest $k$ features given a test input. To measure retrieval performance we use the mean average precision (MAP) score computed as:
\begin{equation}
    \text{MAP} = \frac{\sum_{q=1}^{Q} \text{Ave}(P(q))} {Q}
\end{equation}
over all retrievals. Here, $Q$ is the number of queries and $\text{Ave}(P(q))$ is the average of the precision score for each query $q$.

To justify the effectiveness of the proposed approach, we first performed retrieval on variants of our model and measured the MAP results on the concept patent dataset. When the VAE was first trained on natural images instead of sketches followed by the training of the Siamese framework on top of that model, we achieved an overall MAP of 0.75 for the first 10 retrievals. We also tried training a randomly initialized encoder block for the Siamese network and the MAP dropped to 0.6 for the same retrieval set. Implementing our proposed approach, the overall MAP for the same retrieval set resulted in a MAP score of 0.83.     

\begin{figure}[htb]
  \centering
\includegraphics[width=\linewidth,height=5cm]{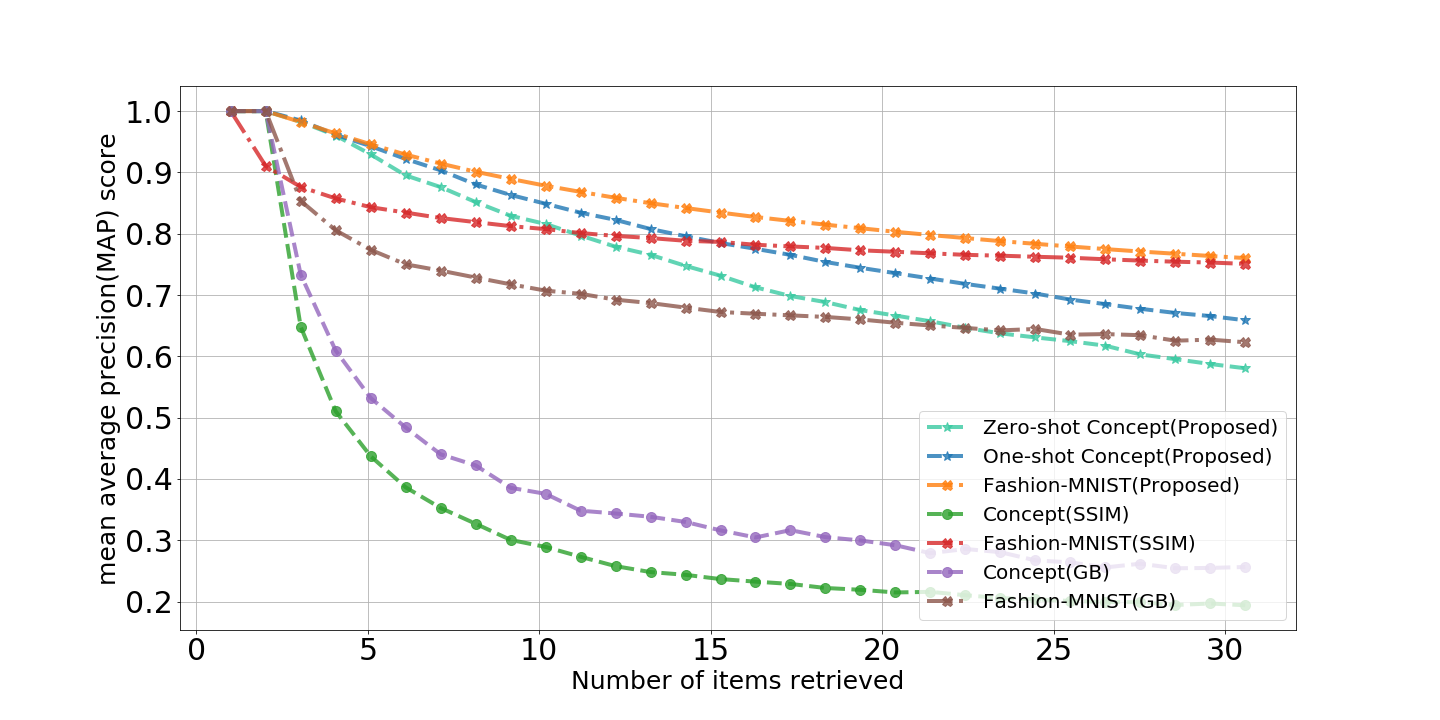}
  \caption{mean average precision(mAP) analysis on Concept Patent dataset(One-Shot and Zero-Shot) and Fashion dataset}
  \label{fig:MAPE}
\end{figure}

For baseline comparisons, we use the standard SSIM and Goldberg(GB) similarity \cite{wong2002image} to compute the structural similarity between all the image pairs on the test data partition and then return the set of $k$-NN images with the highest similarity measure. Table \ref{tab:table_MAP} summarizes the retrieval performance of the proposed framework in comparison to SSIM method for two benchmark datasets: 1) the concept shoe and 2) fashion-MNIST datasets. While SSIM is still used for image matching, classification and retrieval, GB is used as a tool for elastic search in larger datasets\footnote{\href{https://github.com/EdjoLabs/image-match}{https://github.com/EdjoLabs/image-match}}\footnote{\href{https://github.com/dsys/match}{https://github.com/dsys/match}}.  Here, retrieval scores are measured with MAP estimates for 10, 20 and 30 retrieved images per image query. Figure \ref{fig:MAPE} instead shows a more detailed trend of the retrieval performance as a function of the number of retrieved images where our proposed framework outperforms SSIM and GB based retrieval in both datasets. Notice also that our method performance decreases smoothly with increases in items retrieved irrespective of image view-point and intensity variations.

In both Zero-Shot and One-Shot retrieval cases in the concept dataset, performance of the proposed approach is significantly better than SSIM and GB which drops in performance exponentially with increase in the number of retrieved items. Also, note that performance is lower in the Zero-Shot compared to the One-Shot framework caused mainly because of the additional knowledge acquired by the model with regard to the query set in the One-Shot case. Also, note that for fashion-MNIST similar retrieval performances are achieved in both proposed and SSIM methods. This is mainly due to high intra-class similarity, the lack of multiple viewpoints  and the binary scoring used in this case. In contrast, for the concept patent dataset, the similarity scoring which ranges from 0-5 complicates the retrieval process for any other models that are not trained with such a scoring scheme. In both datasets, GB fails to perform well for the retrieval in binary and gray-scale images in contrast to its efficacy for retrieval in large RGB datasets.

\begin{table*} [htb]
	\caption{\label{tab:table_MAP} Retrieval Performances }
	\centering
	\begin{tabular}{lccc}
		\hline
		\\
		
		{Dataset} &{mAP@10} & {mAP@20} & {map@30}  \\
    	\hline

	    \\
	    Concept & \textbf{0.816(ZS on Proposed)} & \textbf{0.667(ZS on Proposed)} & \textbf{0.581(ZS on Proposed)} \\
	    
	    & \textbf{0.842(OS on proposed)} & \textbf{0.721(OS on Proposed)} & \textbf{0.643(OS on Proposed)} \\
	    
	     & 0.348(GB) & 0.280(GB) & 0.247(GB) \\
	      & 0.273(SSIM) & 0.216(SSIM) & 0.193(SSIM) \\
	      
  \hline  
	     	\\
	    Fashion-MNIST & \textbf{0.86(Proposed )} & \textbf{0.81(Proposed)} & \textbf{0.77(Proposed)} \\
	    & 0.807(SSIM) & 0.770(SSIM) & 0.750(SSIM) \\
	    & 0.702(GB) & 0.650(GB) & 0.625(GB) \\
	    \hline

	\end{tabular}
\end{table*}

\section{Conclusion and future work}

We have demonstrated that domain generalization with the edge maps-based inductive  short term learning and latent space fine-tuning based transductive long term learning aids to improve retrieval performance. This two-step process helps to fine tune the feature space by appropriately learning the data manifold. This provides a more meaningful structure of technical diagrams which naive image processing/computer vision techniques are unable to extract. 
Also, we have demonstrated image retrieval using the domain generalization concept on shoe patent images in One-Shot and Zero-Shot settings.  We can extend this framework to other scientific drawings and patent images by pre-training the framework with related datasets.   

To construct the similarity matrix to perform retrieval, we first computed the Euclidean distance/ Cosine similarity between the learning based deep features. These deep features were obtained by transfer learning of a deep learning model for patent images. It was observed that the learned deep features from the supervised classification based task or unsupervised latent representation did not reflect well on the retrieval performance as the learned features were more biased towards the learning objective the framework was trained for. In our approach, we build a pipeline where we fine tune the features obtained with transfer learning with the objective of achieving an improved similarity measure between the features corresponding to different images. This resulted in a better retrieval performance. Also, because of the difficulty involved in quantifying the retrieval performance on training the deep learning model, we instead use a similarity metric implemented via a scoring measure of similarity between image pairs to train the framework. 

In the future, we plan to implement a Bidirectional Generative Adversarial Networks (BiGANs) to learn generative models mapping from simple latent distributions to arbitrarily complex data distributions. This framework would  be able to perform domain generalization across more broad image domains including natural images, sketches, scientific drawings and patent images. 

\section{Acknowledgments}

Research supported by the Laboratory Directed Research and Development program of Los Alamos National Laboratory (LANL) under project number 20200041ER. MB supported by the LANL Applied Machine Learning Summer Research Fellowship(2019).


\end{document}